\documentclass{llncs}
\usepackage{makeidx}

\begin{document} 
\frontmatter

\newtheorem{observation}[theorem]{Observation}
\newtheorem{fact}[theorem]{Fact}
\newtheorem{alg}{Algorithm}
\newenvironment{algorithm}{\fontfamily{cmtt}\selectfont}{\par \bigskip}
\def\poly{{\rm poly}}
\def\cov{{\rm cov}}
\def\var{{\rm var}}
\def\D{{\cal D}}
\def\\KL{{\rm \KL}}
\def\ll{{\cal L}}
\def\R{{\bf R}}
\def\x{{\bf x}}
\def\v{{\bf v}}
\def\q{{\bf q}}
\def\eps{\left({1\over\epsilon}\right)}
\def\del{\left({1\over\delta}\right)}
\def\cost{c}
\def\Cost{C}

\newlength{\boxwidth}
\setlength{\boxwidth}{\linewidth}
\addtolength{\boxwidth}{-1em}

\pagestyle{headings}
\mainmatter

\title{PAC Classification based on PAC Estimates of Label Class
Distributions\thanks{
This work was supported by EPSRC Grant GR/R86188/01.
This work was supported in part by the IST Programme of the
European Community, under the PASCAL Network of Excellence,
IST-2002-506778. This publication only reflects the authors' views.
}}

\author{Nick Palmer\inst{ } \and Paul W. Goldberg\inst{ }}
\authorrunning{Palmer and Goldberg}

\tocauthor{Nick Palmer (University of Warwick),
Paul W. Goldberg (University of Warwick)}

\institute{Dept. of Computer Science, University of Warwick,
Coventry CV4 7AL, U.K.\\
\email{(npalmer|pwg)@dcs.warwick.ac.uk},\\ Research group home page:
\texttt{http://www.dcs.warwick.ac.uk/research/acrg}
}

\maketitle

\begin{abstract}
A standard approach in pattern classification is to estimate the
distributions of the label classes, and then to apply the Bayes
classifier to the estimates of the distributions in order to classify
unlabeled examples. As one might expect, the better our estimates of
the label class distributions, the better the resulting classifier
will be. In this paper we make this observation precise by identifying
risk bounds of a classifier in terms of the quality of the estimates
of the label class distributions. We show how PAC learnability relates
to estimates of the distributions that have a PAC guarantee on their
$L_1$ distance from the true distribution, and we bound the increase
in negative log likelihood risk in terms of PAC bounds on the
KL-divergence.  We give an inefficient but general-purpose smoothing
method for converting an estimated distribution that is good under the
$L_1$ metric into a distribution that is good under the
KL-divergence.
\end{abstract}

{\bf keywords.} Bayes error, Bayes classifier, plug-in decision function

\section{Introduction}

We consider a general approach to pattern classification in which
elements of each class are first used to train a probabilistic model
via some unsupervised learning method. The resulting models for each
class are then used to assign discriminant scores to an unlabeled
instance, and a label is chosen to be the one associated with the
model giving the highest score. For example~\cite{bej-yona} uses this
approach to classify protein sequences, via training a well-known
probabilistic suffix tree model of Ron et al.~\cite{rst:1998} on each
sequence class. Indeed, even where an unsupervised technique is
mainly being used to gain insight into the process that generated two
or more data sets, it is still sometimes instructive to try out the
associated classifier, since the misclassification rate provides a
quantitative measure of the accuracy of the estimated distributions.

The work of~\cite{rst:1998} has led to further related algorithms for
learning classes of probabilistic finite state automata (PDFAs) in
which the objective of learning has been formalized as the estimation
of a true underlying distribution (over strings output by the {\em
target PDFA}) with a distribution represented by a hypothesis
PDFA. The natural discriminant score to assign to a string, is the
probability that the hypothesis would generate that string at random.

As one might expect, the better one's estimates of label class
distributions (the class-conditional densities), the better
should be the associated classifier.
The contribution of this paper is to make precise that observation.
We give bounds on the risk of the associated Bayes
classifier\footnote{The {\em Bayes classifier} associated with two or
more probability distributions is the function that maps an element $x$
of the domain to the label associated with the probability distribution
whose value at $x$ is largest. This is of course a well-known approach
for classification, see~\cite{Duda:1973}.}
in terms of the quality of the estimated distributions.

These results are partly motivated by our interest in the relative
merits of estimating a class-conditional
distribution using the variation distance, as
opposed to the KL-divergence (defined in the next
section). In~\cite{Clark:2004} it has been shown how to learn a class
of PDFAs using KL-divergence, in time polynomial in a set of
parameters that includes the expected length of strings output by the
automaton. In~\cite{Palmer:2005} we show how to learn this class with
respect to variation distance, with a polynomial sample-size bound
that is independent of the length of output strings. Furthermore, it
can be shown that it is necessary to switch to the weaker criterion of
variation distance, in order to achieve this. We show here that this
leads to a different---but still useful---performance guarantee
for the Bayes classifier.

Abe and Warmuth~\cite{aw:1992} study the problem of learning
probability distributions using the KL-divergence, via classes of
probabilistic automata. Their criterion for learnability is that---for
an unrestricted input distribution $D$---the hypothesis PDFA
should be almost (i.e.~within $\epsilon$) as close as possible to $D$.
Abe, Takeuchi and Warmuth~\cite{Abe:2001} study the negative
log-likelihood loss function in the context of learning {\em
stochastic rules}, i.e.~rules that associate an element of the domain
$X$ to a probability distribution over the range $Y$. We show here
that if two or more label class distributions are learnable in the sense
of~\cite{aw:1992}, then the resulting stochastic rule (the conditional
distribution over $Y$ given $x\in X$) is learnable in the sense
of~\cite{Abe:2001}.

We show that if instead the label class distributions are well
estimated using the variation distance, then the associated classifier
may not have a good negative log likelihood risk, but will have a {\em
misclassification rate} that is close to optimal. This result is for
general $k$-class classification, where distributions may overlap
(i.e. the optimum misclassification rate may be positive).  We also
incorporate variable misclassification penalties (sometimes one might
wish a false positive to cost more than a false negative), and show
that this more general loss function is still approximately minimized
provided that discriminant likelihood scores are rescaled appropriately.

As a result we show that PAC-learnability and more generally
p-concept\footnote{p-concepts are functions probabilistically mapping
elements of the domain to 2 classes.} learnability~\cite{Kearns:1993},
follows from the ability to learn class distributions in the setting
of Kearns et al.~\cite{Kearns:1994}.  Papers such
as~\cite{Cryan:2001,MR:2005,fos:2005} study the problem of learning
various classes of probability distributions with respect to
KL-divergence and variation distance, in this setting.

It is well-known (noted in~\cite{Kearns:1993}) that learnability with
respect to KL-divergence is stronger than learnability with respect to
variation distance.  Furthermore, the KL-divergence is usually used
(for example in~\cite{Clark:2004,Hoffgen:1993}) due to the property
that when minimized with respect to an sample, the empirical
likelihood of that sample is maximized.  An algorithm that learns with
respect to variation distance can sometimes be converted to one that
learns with respect to KL-divergence by a smoothing
technique~\cite{Cryan:2001}, when the domain is $\{0,1\}^n$, and $n$
is a parameter of the learning problem.  In this paper we give a
related smoothing rule that applies to the version of the PDFA
learning problem where we seem to ``need'' to use the variation
distance. However, the smoothed distribution does not have an
efficient representation, and requires the probabilities used in the
target PDFA to have limited precision.

\subsection{Notation and Terminology}
\label{notation}

In $k$-class classification, labeled examples are generated by
distribution $D$ over $X \times \{1,...,k\}$.  We consider the problem
of predicting the label $\ell$ associated with $x \in X$, where $x$ is
generated by the marginal distribution of $D$ on $X$, $D|_X$.  A
non-negative cost is incurred for each classification, based either on
a cost matrix (where the cost depends upon both the hypothesized label
and the true label) or the negative log-likelihood of the true label
being assigned. The aim is to optimize the expected cost given by the
occurrence of a randomly generated example.  We refer to the expected
cost associated with any classifier $f:X \rightarrow \{1,...,k\}$, as
{\em risk} (as described by Vapnik~\cite{Vapnik:2000}), denoted as $R(f)$.

Let $D_{\ell}$ be $D$ restricted to points $(x,\ell)$, $\ell =
\{1,...,k\}$.  $D$ is a mixture $\sum_{\ell = 1}^k g_\ell D_\ell$,
where $\sum_{i=1}^k g_i = 1$, and $g_\ell$ is the {\em class prior} of
class $\ell$---the probability that a randomly generated data point
has label $\ell$.

In Section~\ref{results} it is shown that if we have upper bounds on
the inaccuracy of the estimated distributions of each class label,
then we can derive bounds on the risk associated with the
classifiers. Suppose $D$ and $D'$ are probability distributions over
the same domain $X$.  We define the $L_1$ distance as $L_1(D,D') =
\int_X \vert D(x) - D'(x)\vert\ dx$.  We usually assume that $X$ is a
discrete domain, in which case
\[
L_1(D,D') = \sum_{x \in X} \vert D(x) - D'(x)\vert.
\]
The KL-divergence from $D$ to $D'$ is defined as
\[
I(D || D') = \sum_{x\in X}D(x)\log\left({{D(x)}\over{D'(x)}}\right).
\]

\subsection{Learning Framework}
\label{framework}

In the PAC-learning framework an algorithm receives labeled
samples generated independently according to distribution $D$ over
$X$, where distribution $D$ is unknown, and where labels are
generated by an unknown function $f$ from a known class of functions
${\cal F}$. The algorithm must output a hypothesis $h$ from a class
of hypotheses ${\cal H}$, such that with probability at least $1 -
\delta$, $err_h \leq \epsilon$, where $\epsilon$ and $\delta$ are
parameters. Notice that in this setting, if $f \in {\cal H}$, then
$err^* = 0$, where $err^*$ is the error associated with the optimal
hypothesis.

We use a variation on the framework used in~\cite{Kearns:1993} for
learning p-concepts, which adopts performance measures from the PAC
model, extending this to learn stochastic rules with $k$ classes.
Therefore it is the case that $err^* = \inf_{h\in {\cal H}}\{err_h\}$.
The aim of the learning algorithm in this framework is to output a
hypothesis $h\in \cal H$ such that with probability of at least
$1-\delta$, the error $err_h$ of $h$ satisfies $err_h \leq err^* +
\epsilon$.

Our notion of learning distributions is similar to that of Kearns et
al.~\cite{Kearns:1994}.

\begin{definition}
\label{efficient_learning} Let ${\cal D}_n$ be a class of
distributions.  ${\cal D}_n$ is said to be efficiently learnable if
an algorithm $A$ exists, such that given $\epsilon>0$ and $\delta>0$
and access to randomly drawn examples (see below) from any unknown
target distribution $D\in {\cal D}_n$, $A$ runs in time polynomial
in $\eps$, $\del$ and $n$ and returns a probability distribution $D'$
that with probability at least $1-\delta$ is within $L_1$-distance
(alternatively KL-divergence) $\epsilon$ of $D$.
\end{definition}

We define p-concepts as introduced by Kearns and
Shapire~\cite{Kearns:1993}. This definition is for 2-class
classification, but generalizes in a natural way to more than 2
classes.

\begin{definition}
\label{pconcept_learning} A Probabilistic Concept (or p-concept) $f$ on
domain $X$ is given by a real-valued function $p_f:X \rightarrow
[0,1]$. An observation of $f$ consists of some $x\in X$ together
with a 0/1 label $\ell$ with $\Pr(\ell=1)=p_f(x)$.
\end{definition}




\section{Results}
\label{results}

In Section~\ref{risk_bounds} we give bounds on the risk associated
with a hypothesis, with respect to the accuracy of the approximation
of the underlying distribution generating the instances.  In
Section~\ref{lowerbounds} we show that these bounds are close to
optimal, and in Section~\ref{paclearnability} we give corollaries
showing what these bounds mean for PAC learnability.

We define the accuracy of an approximate distribution in
terms of $L_1$ distance and KL divergence, both of which are
commonly used measurements.  It is assumed that the class priors of
each class label are known.

\subsection{Bounds on Increase in Risk}
\label{risk_bounds}

First we examine the case where the accuracy of the hypothesis
distribution is such that the distribution for each class label is
within $L_1$ distance $\epsilon$ of the true distribution for that
label, for some $0 \leq \epsilon \leq 1$.  A cost matrix $\Cost$ specifies
the cost associated with any classification, where the cost of
classifying a data point which has label $i$ as some label $j$ is
denoted as $\cost_{ij}$ (where $\cost_{ij}\geq 0$).  It is usually the
case that $\cost_{ij}=0$ for $i=j$. We introduce the following
notation:

Given classifier $f$ over discrete domain $X$, $f:X \rightarrow
\{1,...,k\}$, the risk of $f$ is given by
\[
R(f) = \sum_{x \in X}  \sum^{k}_{i=1} \cost_{if(x)}.g_i.D_i(x).
\]

Let $f^*$ be the Bayes optimal classifier, i.e.~the function with the
minimal risk, or optimal expected cost, and $f'(x)$ is the function
with optimal expected cost with respect to alternative distributions
$D'_i, i\in \{1,...,k\}$. For $x \in X$,
\[
\begin{array}{rcl}
f^*(x) & = & \arg \min_j\sum^{k}_{i=1}\cost_{ij}.g_i.D_i(x)\\
f'(x)  & = & \arg\ \min_j\sum^{k}_{i=1}\cost_{ij}.g_i.D'_i(x).
\end{array}
\]

\begin{theorem}\footnote{This result is essentially a generalization
of Exercise 2.10 of Devroye et al's textbook~\cite{Devroye}, from
2 class to multiple classes, and in addition we show here that variable
misclassification costs can be incorporated. This is the closest
thing we have found to this Theorem that has already appeared, but
we suspect that other related results may have appeared. We would welcome
any further information or references on this topic.
Theorem~\ref{upper-bound_kl-divergence} is another result which
we suspect may be known, but likewise we have found no statement of it.}
\label{upper-bound_variation_distance}
Let $f^*$ be the Bayes optimal classifier and let $f'$ be the
Bayes classifier associated with estimated distributions $D'_i$.
Suppose that for each label
$i\in\{1,...,k\}$, $L_1(D_i,D'_i) \leq \epsilon/g_i$.
Then $R(f') \leq R(f^*) + \epsilon.k.\max_{ij}\{a_{ij}\}.$
\end{theorem}

\begin{proof}
Let $R_f(x)$ be the contribution from $x\in X$ towards the total
expected cost associated with classifier $f$.  For $f$ such that
$f(x) = j$,
\[
R_f(x) = \sum^{k}_{i=1} \cost_{ij}.g_i.D_i(x).
\]

Let $\tau_{\ell'-\ell}(x)$ be the increase in risk for labelling $x$
as $\ell'$ instead of $\ell$, so that

\begin{equation}\label{tau}
\begin{array}{rcl}
\tau_{\ell'-\ell}(x) & = & \sum^{k}_{i=1} \cost_{i\ell'}.g_i.D_i(x) -
\sum^{k}_{i=1} \cost_{i \ell}.g_i.D_i(x) \\
                     & = & \sum^{k}_{i=1}
(\cost_{i\ell'}-\cost_{i\ell}).g_i.D_i(x).
\end{array}
\end{equation}

Note that due to the optimality of $f^*$ on $D_i$, $\forall x \in
X:\tau_{f'(x)-f^*(x)}(x) \geq 0$.  In a similar way, the expected
contribution to the total cost of $f'$ from $x$ must be less than or
equal to that of $f^*$ with respect to $D'_i$ -- given that $f'$ is
chosen to be optimal on the $D'_i$ values.  We have:

\[
\sum^{k}_{i=1}\cost_{i f'(x)}.g_i.D'_i(x) \leq \sum^{k}_{i=1}
\cost_{if^*(x)}.g_i.D'_i(x).
\]

Rearranging, we have
\begin{equation}\label{0001}
\sum^{k}_{i=1}D'_i(x).g_i.\left(\cost_{if^*(x)} -
\cost_{i f'(x)}\right) \geq 0.
\end{equation}

  From~(\ref{tau}) and~(\ref{0001}) it can be seen that
\[
\begin{array}{rl}
\tau_{f'(x)-f^*(x)}(x) & \leq \left(D_i-D'_i(x)\right).
g_i.\left(\cost_{i f'(x)}-\cost_{i f^*(x)}\right)\\
& \leq \sum_{i=1}^k
\left|\left(D_i-D'_i(x)\right)\right|.g_i.\left|
\left(\cost_{i f'(x)}-\cost_{i f^*(x)}\right)\right|.
\end{array}
\]

Let $d_i(x)$ be the difference between the probability densities of
$D_i$ and $D'_i$ at $x \in X$, $d_i(x) = \left|D_i(x) -
D'_i(x)\right|$.  Therefore,

\[
\tau_{f'(x)-f^*(x)}(x) \leq \sum^{k}_{i=1} |\cost_{i f'(x)} - \cost_{i
f^*(x)}|.g_i.d_i(x) \leq \tau_{f'(x)-f^*(x)}(x) \leq \sum^{k}_{i=1}
\max_j\{\cost_{ij}\}.g_i.d_i(x).
\]

In order to bound the expected cost, it is necessary to sum over the
range of $x \in X$:

\begin{equation}\label{0003}
\sum_{x\in X} \tau_{f'(x)-f^*(x)}(x) \leq \sum_{x\in X}
\sum^{k}_{i=1} \max_j\{\cost_{i j}\}.g_i.d_i(x)= \sum^{k}_{i=1}
\max_j\{\cost_{i j}\}.g_i.\sum_{x\in X} d_i(x).
\end{equation}

Since $L_1(D_i,D'_i)\leq \epsilon/g_i$ for all $i$, ie. $\sum_{x
\in X} d_i(x) \leq \epsilon/g_i$, it follows from~(\ref{0003})
that
\[
\sum_{x\in X} \tau(x) \leq \sum^{k}_{i=1} \max_j\{\cost_{i
j}\}.g_i.\left(\frac{\epsilon}{g_i}\right).
\]

This expression gives an upper bound on expected cost for labelling
$x$ as $f'(x)$ instead of $f^*(x)$.  By definition,
\[
\sum_{x \in X} \tau(x) = R(f') - R(f^*).
\]

Therefore it has been shown that
\[
R(f') \leq R(f^*) + \epsilon.\sum^{k}_{i=1} \max_j\{\cost_{ij}\} \leq
R(f^*) + \epsilon.k.\max_{ij}\{\cost_{ij}\}.
\]
\qed
\end{proof}

We next prove a corresponding result in terms of KL-divergence, which
uses the negative log-likelihood of the correct label as the cost
function. We define $\Pr_i(x)$ to be the probability that a data point
at $x$ has label $i$, such that $\Pr_i(x) =
g_i.D_i(x)\left(\sum_{j=1}^k g_j.D_j(x)\right)^{-1}$.  Given a
function $f : X \rightarrow \R^k$, where $f(x)$ is a prediction of the
probabilities of $x$ having each label $i \in \{1,...,k\}$ (so
$\sum^{k}_{i = 1}f_i(x) =1$), the risk associated with $f$ can be
expressed as
\begin{equation}\label{eqn:llrisk}
R(f) = \sum_{x \in X} D(x)
\sum^{k}_{i=1}-\log(f_i(x)).{\Pr}_i(x).
\end{equation}

Let $f^*:X\longrightarrow {\bf R}^k$ output the true class label
distribution for an element of $X$. From Equation~(\ref{eqn:llrisk})
it can be seen that
\begin{equation}
\label{kl_001} 
R(f^*) = \sum_{x \in X} D(x) \sum^{k}_{i=1}
-\log({\Pr}_i(x)).{\Pr}_i(x).
\end{equation}

\begin{theorem}
\label{upper-bound_kl-divergence}
For $f:X\longrightarrow {\bf R}^k$ suppose that $R(f)$ is given
by~(\ref{eqn:llrisk}). If for each label $i \in
\{1,...,k\}$, $I(D_{i} || D'_{i}) \leq \epsilon/g_i$,
then $R(f') \leq R(f^*) + k\epsilon$.
\end{theorem}
\begin{proof}
Let $R_f(x)$ be the contribution at $x \in X$ to the risk associated
with classifier $f$, $R_f(x)=\sum_{i=1}^k-\log(f_i(x)).\Pr_i(x)$.
Therefore $R(f')=\sum_{x\in X}D(x).R_{f'}(x)$.

We define $\Pr'_i(x)$ to be the estimated probability that a data
point at $x\in X$ has label $i\in \{1,...,k\}$, from distributions
$D'_i$, such that
$\Pr'_i(x)=g_i.D'_i\left(\sum_{j=1}^k
 g_j.D'_j(x)\right)^{-1}$.

\[
R_{f'}(x) = D(x).\sum^{k}_{i = 1}-\log\left({\Pr}'_i(x)\right).{\Pr}_i(x).
\]

Let $\xi(x)$ denote the contribution to additional risk incurred
from using $f'$ as opposed to $f^*$ at $x \in X$. From
(\ref{kl_001}) it can be seen that

\begin{eqnarray*}
\xi(x) & = & R_{f'}(x) - D(x).\sum^{k}_{i =
1}-\log\left({\Pr}_i(x)\right).{\Pr}_i(x)\\
  & = & D(x).\sum^{k}_{i = 1}{\Pr}_i(x).\left(\log\left({\Pr}_i(x)\right)
  -\log\left({\Pr}'_{i}(x)\right)\right)\\
  & = & D(x).\sum^{k}_{i = 1}\left(\frac{g_i.D_i(x)}{\sum_{j=1}^k g_j.D_j(x)}\right)\left(\log\left(\frac{g_i.D_i(x)}{\sum_{j=1}^k g_j.D_j(x)}\right)
  -\log\left(\frac{g_i.D'_i(x)}{\sum_{j=1}^k g_j.D'_j(x)}\right)\right)\\
  & = & D(x).\sum^{k}_{i=1}\left(\left(\frac{g_i.D_{i}(x)}{\sum^{k}_{j
=1}g_j.D_j(x)}\right).\left(\log\left(\frac{g_i.D_{i}(x)}{g_i.D'_{i}(x)}\right)
- \log\left(\frac{\sum^{k}_{j=1}g_j.D_j(x)}{\sum^{k}_{j=1}g_j.D'_j(x)}\right)\right)\right).
\end{eqnarray*}


We define \(D'\) such that $D'(x) = \sum^{k}_{i = 1}
g_i.D'_{i}(x)$.  Since it is the case that $D(x) = \sum^{k}_{i
= 1} g_i.D_{i}(x)$, $\xi(x)$ can be rewritten as

\[
\begin{array}{rl}
\xi\left(x\right) &= D(x).\sum^{k}_{i =
1}\left(\frac{g_i.D_{i}(x)}{D(x)}\right).\left(\log\left(\frac{g_i.D_{i}(x)}{g_i.D'_{i}(x)}\right)
- \log\left(\frac{D(x)}{D'(x)}\right)\right)\\
&= \sum^{k}_{i =
1}\left(g_i.D_{i}(x)\log\left(\frac{D_{i}(x)}{D'_{i}(x)}\right)\right)
- D(x)\log\left({D(x) \over D'(x)}\right).
\end{array}
\]

We define $I(D||D')(x)$ to be the contribution at $x \in X$ to the
KL-divergence, such that
$I(D||D')(x)=D(x)\log\left(D(x)/D'(x)\right)$.  It follows that

\begin{equation}
\label{sumofxi}
\sum_{x \in X} \xi(x) = \sum^{k}_{i = 1}\left(
g_i.I(D_{i}||D'_{i})\right) - I(D||D').
\end{equation}

We know that the KL divergence between $D_i$ and $D'_i$ is
bounded by $\epsilon/g_i$ for each label $i\in\{1,...,k\}$,
so (\ref{sumofxi}) can be rewritten as

\[
\sum_{x \in X} \xi(x) \leq \sum^{k}_{i = 1}\left(
g_i.\left(\frac{\epsilon}{g_i}\right)\right) - I(D||D')
\leq k.\epsilon - I(D||D').
\]

Due to the fact that the KL-divergence between two distributions is
non-negative, an upper bound on the cost can be obtained by letting
$I(D||D') = 0$, so $R(f') - R(f^*) \leq k\epsilon$.  Therefore it has
been proved that $R(\hat{f}) \leq R(f^*) + k\epsilon$.\qed
\end{proof}

\subsection{Lower Bounds}
\label{lowerbounds}

In this section we give lower bounds corresponding to the two upper
bounds given in Section~\ref{results}.

\begin{example}
\label{ex001} Consider a distribution $D$ over domain $X =
\{x_0,x_1\}$, from which data is generated with labels $0$ and $1$
and there is an equal probability of each label being generated
($g_0 = g_1 = \frac{1}{2}$).
 $D_i(x)$ denotes the probability that a point is generated at $x\in
X$ given that it has label $i$.  $D_0$ and $D_1$ are distributions
over $X$, such that at $x\in X$, $D(x) = \frac{1}{2}(D_0(x) +
D_1(x))$.

Suppose that $D'_0$ and $D'_1$ are approximations of $D_0$
and $D_1$, and that
$L_1(D_0,D'_0)=\frac{\epsilon}{g_0}=2\epsilon$ and
$L_1(D_1,D'_1)=\frac{\epsilon}{g_1}=2\epsilon$, where $\epsilon
= \epsilon' + \gamma$ (and $\gamma$ is an arbitrarily small
constant).

Given the following distributions, assuming that a misclassification
results in a cost of $1$ and that a correct classification results
in no cost, it can be seen that $R(f^*) = \frac{1}{2}-\epsilon'$:

\[D_0(x_0)=\frac{1}{2}+\epsilon',
D_0(x_1)=\frac{1}{2}-\epsilon',\]
\[D_1(x_0)=\frac{1}{2}-\epsilon',
D_1(x_1)=\frac{1}{2}+\epsilon'.\]

Now if we have approximations $D'_0$ and $D'_1$ as shown
below, it can be seen that $f'$ will misclassify for every value
of $x\in X$:

\[D'_0(x_0)=\frac{1}{2} - \gamma,   D'_0(x_1)=\frac{1}{2} + \gamma,\]
\[D'_1(x_0)=\frac{1}{2} + \gamma,   D'_1(x_1)=\frac{1}{2} - \gamma.\]

This results in $R(f')=\frac{1}{2}+\epsilon'$. Therefore
$R(f')=R(f^*)+2\epsilon'=R(f^*)+2(\epsilon-\gamma)$.

\end{example}

In this example the risk is only $2\gamma$ under $R(f^*) +
\epsilon.k.\max_j\{a_{ij}\}$, since $k=2$.  A similar example can be
used to give upper bounds to the lower bound given in
Theorem~\ref{upper-bound_kl-divergence}.

\begin{example}
\label{ex002}
Consider distributions $D_0$, $D_1$, $D'_0$ and $D'_1$ over
domain $X = \{x_0,x_1\}$ as defined in Example~\ref{ex001}.  It can be
seen that the KL-divergence between each label's distribution and its
approximated distribution is

\[I(D_0||D'_0)=I(D_1||D'_1)=\left(\frac{1}{2}+\epsilon'\right)\log\left(\frac{\frac{1}{2}
+ \epsilon'}{\frac{1}{2}-\gamma}\right) +
\left(\frac{1}{2}-\epsilon'\right)\log\left(\frac{\frac{1}{2} -
\epsilon'}{\frac{1}{2} + \gamma}\right).
\]

The optimal risk, measured in terms of negative log-likelihood, can be
expressed as
$R(f^*)=-\left(\frac{1}{2}+\epsilon'\right)\log\left(\frac{1}{2}+\epsilon'\right)-\left(\frac{1}{2}-\epsilon'\right)\log\left(\frac{1}{2}-\epsilon'\right)$.
The risk incurred by using $f'$ as the discriminant function is
$R(f')=-\left(\frac{1}{2}+\epsilon'\right)\log\left(\frac{1}{2}-\gamma\right)-\left(\frac{1}{2}-\epsilon'\right)\log\left(\frac{1}{2}+\gamma\right)$.
Therefore,

\[R(f') = R(f^*)+\left(\frac{1}{2}
+\epsilon'\right)\log\left(\frac{\frac{1}{2}
+\epsilon'}{\frac{1}{2}-\gamma}\right)+\left(\frac{1}{2}
-\epsilon'\right)\log\left(\frac{\frac{1}{2} -
\epsilon'}{\frac{1}{2}+\gamma}\right) = R(f^*)+\epsilon.
\]

\end{example}

\subsection{Learning near-optimal classifiers in the PAC sense}
\label{paclearnability}

We show that the results of Section~\ref{risk_bounds} imply
learnability within the framework defined in Section~\ref{framework}.

The following corollaries refer to algorithms $A_{class}$ and
$A_{class'}$.  These algorithms generate classifier functions
$f':X\longrightarrow\{1,2,\ldots,k\}$,
which label data in a $k$-label classification problem, using $L_1$
distance and $KL$-divergence respectively as measurements of accuracy.

Corollary~\ref{cor_l1} shows (using
Theorem~\ref{upper-bound_variation_distance}) that a near optimal
classifier can be constructed given that an algorithm exists which
approximates a distribution over positive data in polynomial time.  We
are given cost matrix $\Cost$, and assume knowledge of the class priors
$g_i$.

\begin{corollary}
\label{cor_l1}
If an algorithm $A_{L_1}$ approximates distributions within $L_1$
distance $\epsilon'$ with probability at least $1-\delta'$, in time
polynomial in $1/\epsilon'$ and $1/\delta'$, then an algorithm
$A_{class}$ exists which (with probability $1-\delta$) generates a
discriminant function $f'$ with an associated risk of at most
$R(f^*)+\epsilon$, and $A_{class}$ is polynomial in $1/\delta$ and
$1/\epsilon$.
\end{corollary}

\begin{proof}
$A_{class}$ is a classification algorithm which uses
unsupervised learners to fit a distribution to each label
$i\in\{1,...,k\}$, and then uses the Bayes classifier with respect
to these estimated distributions, to label data.

$A_{L_1}$ is a PAC algorithm which learns from a sample of positive
data to estimate a distribution over that data. $A_{class}$ generates
a sample $N$ of data, and divides $N$ into sets $\{N_1,...,N_k\}$,
such that $N_i$ contains all members of $N$ with label $i$.  Note that
for all labels $i$, $|N_i|\approx g_i.|N|$.

With a probability of at least $1-\frac{1}{2}(\delta/k)$, $A_{L_1}$
generates an estimate $D'$ of the distribution $D_i$ over label $i$,
such that $L_1(D_i,D') \leq
\epsilon\left(g_i.k.\max_{ij}\{\cost_{ij}\}\right)^{-1}$. Therefore the
size of the sample $|N_i|$ must be polynomial in
$g_i.k.\max_{ij}\{\cost_{ij}\}/\epsilon$ and $k/\delta)$.  For all
$i\in\{1,...,k\}$ $g_i \leq 1$, so $|N_i|$ is polynomial in
$\max_{ij}\{\cost_{ij}\}$, $k$, $1/\epsilon$ and $1/\delta$.

When $A_{class}$ combines the distributions returned by the $k$
iterations of $A_{L_1}$, there is a probability of at least
$1-\delta/2$ that all of the distributions are within
$\epsilon\left(g_i.k.\max_{ij}\{\cost_{ij}\}\right)^{-1}$ $L_1$ distance
of the true distributions (given that each iteration received a
sufficiently large sample). We allow a probability of $\delta/2$ that
the initial sample $N$ did not contain a good representation of all
labels ($\neg\forall i\in\{1,...k\}:|N_i| \approx g_i.|N|$), and as
such -- one or more iteration of $A_{L_1}$ may not have received a
sufficiently large sample to learn the distribution accurately.

Therefore with probability at least $1-\delta$, all approximated
distributions are within $\epsilon(g_i.k.\max_{ij}\{\cost_{ij}\})^{-1}$
$L_1$ distance of the true distributions.  If we use the classifier
which is optimal on these approximated distributions, $f'$, then the
increase in risk associated with using $f'$ instead of the Bayes
Optimal Classifier, $f^*$, is at most $\epsilon$. It has been shown
that $A_{L_1}$ requires a sample of size polynomial in $1/\epsilon$,
$1/\delta$, $k$ and $\max_{ij}\{\cost_{ij}\}$. It follows that
\[
|N| = \sum^k_{i=1} |N_i| = \sum^k_{i=1}
p\left(\frac{1}{\epsilon},\frac{1}{\delta},k,
\max_{ij}\{\cost_{ij}\}\right)\in
O\left(p\left(\frac{1}{\epsilon},
\frac{1}{\delta},k,\max_{ij}\{\cost_{ij}\}\right)\right).
\]
\qed
\end{proof}

Corollary~\ref{cor_kl} shows (using
Theorem~\ref{upper-bound_kl-divergence}) how a near optimal classifier
can be constructed given that an algorithm exists which approximates a
distribution over positive data in polynomial time.

\begin{corollary}
\label{cor_kl}
If an algorithm $A_{KL}$ has a probability of at least $1-\delta$ of
approximating distributions within $\epsilon$ $KL$-divergence, in time
polynomial in $1/\epsilon$ and $1/\delta$, then an algorithm
$A_{class'}$ exists which (with probability $1-\delta$) generates a
function $f'$ that maps $x\in X$ to a conditional distribution
over class labels of $x$, with an associated log-likelihood risk of
at most $R(f^*)+\epsilon$, and $A_{class'}$ is polynomial in
$1/\delta$ and $1/\epsilon$.
\end{corollary}

\begin{proof}
$A_{class'}$ is a classification algorithm using the same
method as $A_{class}$ in Corollary~\ref{cor_l1}, whereby a
sample $N$ is divided into sets $\{N_1,...,N_k\}$, and each set is
passed to algorithm $A_{KL}$ where a distribution is
estimated over the data in the set.

With a probability of at least $1 - \frac{1}{2}(\delta/k)$,
$A_{KL}$ generates an estimate $D'$ of the distribution
$D_i$ over label $i$, such that $I(D_i || D') \leq
\epsilon(g_i.k)^{-1}$. Therefore the size of the
sample $|N_i|$ must be polynomial in
$g_i.k/\epsilon$ and
$k/\delta$.  Since $g_i \leq 1$, $|N_i|$ is
polynomial in $k/\epsilon$ and
$k/\delta$.

When $A_{class'}$ combines the distributions returned by the
$k$ iterations of $A_{KL}$, there is a probability of at
least $1-\delta/2$ that all of the distributions are within
$\epsilon(g_i.k)^{-1}$ $KL$-divergence of the true
distributions. We allow a probability of $\delta/2$ that the
initial sample $N$ did not contain a good representation of all
labels ($\neg\forall i\in\{1,...k\}:|N_i| \approx g_i.|N|$).

Therefore with probability at least $1-\delta$, all approximated
distributions are within $\epsilon(g_i.k)^{-1}$ $KL$-divergence of the
true distributions.  If we use the classifier which is optimal on
these approximated distributions, $f'$, then the increase in risk
associated with using $f'$ instead of the Bayes Optimal Classifier
$f^*$, is at most $\epsilon$. It has been shown that $A_{KL}$ requires
a sample of size polynomial in $1/\epsilon$, $1/\delta$ and $k$.
Let $p(1/\epsilon,1/\delta)$ be an upper bound on the time and
sample size used by $A_{KL}$. It follows that
\[
|N| = \sum^k_{i=1} |N_i| = \sum^k_{i=1}
p\left(\frac{1}{\epsilon},\frac{1}{\delta}\right) \in
O\left(k.p\left(\frac{1}{\epsilon},\frac{1}{\delta}\right)\right).
\]
\qed
\end{proof}

\subsection{Smoothing: from $L_1$ distance to KL-divergence}

Given a distribution that has accuracy $\epsilon$ under the $L_1$
distance, is there a generic way to ``smooth'' it so that it has
similar accuracy under the KL-divergence? From~\cite{Cryan:2001} this
can be done for $X=\{0,1\}^n$, if we are interested in algorithms that
are polynomial in $n$ in addition to other parameters. Suppose however
that the domain is bit strings of unlimited length. Here we
give a related but weaker result in terms of bit strings that are
used to represent distributions, as opposed to members of the domain.
We define class ${\cal D}$ of distributions specified by bit strings,
such that each member of $\cal D$ is a distribution on discrete domain
$X$, represented by a discrete probability scale.  Let $L_D$ be the
length of the bit string describing distribution $D$.
Note that there are at most $2^{L_D}$ distributions
in $\cal D$ represented by strings of length $L_D$.

\begin{lemma}
\label{smoothing}
Suppose $D\in\cal D$ is learnable under $L_1$ distance in time
polynomial in $\delta$, $\epsilon$ and $L_D$. Then $\cal D$ is
learnable under KL-divergence, with polynomial sample size.
\end{lemma}

\begin{proof}
Let $D$ be a member of class ${\cal D}$, represented by a bit string
of length $L_D$, and let algorithm $A$ be an algorithm which takes an
input set $S$ (where $|S|$ is polynomial in $\epsilon$, $\delta$ and
$L_D$) of samples generated i.i.d. from distribution $D$, and with
probability at least $1-\delta$ returns a distribution $D_{L_1}$, such
that $L_1(D,D_{L_1})\leq \epsilon$.

Let $\xi = \frac{1}{12}\left(\epsilon^2/L_D\right)$.  We define
algorithm $A'$ such that with probability at least $1-\delta$, $A'$
returns distribution $D'_{L_1}$, where $L_1(D,D'_{L_1})\leq \xi$.
Algorithm $A'$ runs $A$ with sample $S'$, where $|S'|$ is polynomial
in $\xi$, $\delta$ and $L_D$ (and it should be noted that $|S'|$ is
polynomial in $\epsilon$, $\delta$ and $L_D$).

We define $D_{L_D}$ to be the unweighted mixture of all distributions
in $\cal D$ represented by length $L_D$ bit strings, $D_{L_D}(x) =
2^{-L_D}\sum_{D\in\cal D}D(x)$.  We now define distribution $D'_{KL}$
such that $D'_{KL}(x)=(1-\xi)D'_{L_1}(x)+\xi.D_{L_D}(x)$.

By the definition of $D'_{KL}$, $L_1(D'_{L_1},D'_{KL})\leq 2\xi$.
With probability at least $1-\delta$, $L_1(D,D'_{L1})\leq \xi$, and
therefore with probability at least $1-\delta$, $L_1(D,D'_{KL})\leq
3\xi$.

We define $X_<=\{x\in X|D'_{KL}(x) < D(x)\}$.  Members of $X_<$
contribute positively to $I(D||D'_{KL})$.  Therefore

\begin{equation}\label{1aa}
\begin{array}{rcl}
I(D||D'_{KL}) & \leq & \sum_{x\in
X_<}D(x)\left(\frac{\log(D(x))}{\log(D'_{KL}(x))}\right) \\
& = & \sum_{x\in X_<}(D(x)-D'_{KL}(x))\left(\frac{\log(D(x))}
{\log(D'_{KL}(x))}\right) \\
&   & + \sum_{x\in
X_<}D'_{KL}(x)\left(\frac{\log(D(x))}{\log(D'_{KL}(x))}\right).
\end{array}
\end{equation}

We have shown that $L_1(D,D'_{KL})\leq 3\xi$, so $\sum_{x\in
X_<}(D(x)-D'_{KL}(x))\leq 3\xi$.  Analysing the first term
in~(\ref{1aa}),

\[
\sum_{x\in X_<}(D(x)-D'_{KL}(x))\left(\frac{\log(D(x))}
{\log(D'_{KL}(x))}\right) \leq
3\xi\max_{x\in X_<}\left(\frac{\log(D(x))}{\log(D'_{KL}(x))}\right).
\]

Note that for all $x\in X$, $D'_{KL}(x)\geq \xi.2^{-L_D}$.
It follows that
\[
\max_{x\in X_<}\left(\frac{\log(D(x))}{\log(D'_{KL}(x))}\right)
\leq \log (2^{L_D}/\xi)
= L_D-\log(\xi).
\]

Examining the second term in (\ref{1aa}),
\[
\sum_{x\in X_<}D'_{KL}(x)\left(\frac{\log(D(x))}{\log(D'_{KL}(x))}\right) =
\sum_{x\in X_<}D'_{KL}(x)\left(\frac{\log(D'_{KL}(x)+h_x)}{\log(D'_{KL}(x))}\right),
\]
where $h_x = D(x)-D'_{KL}(x)$, which is a positive quantity for all
$x\in X_<$.  Due to the concavity of the logarithm function, it
follows that
\[
\begin{array}{rcl}
\sum_{x\in X_<}D'_{KL}(x)\left(\frac{\log(D'_{KL}(x)+h_x)}
{\log(D'_{KL}(x))}\right) & \leq &
\sum_{x\in X_<}D'_{KL}(x)h_x
\left[\frac{d}{dy}(\log(y))\right]_{y=D'_{KL}(x)} \\
& = & \sum_{x\in X_<}h_x \leq 3\xi.
\end{array}
\]

Therefore, $I(D||D'_{KL}) \leq 3\xi(1 + L_D -\log(\xi))$.  For values
of $\xi \leq \frac{1}{12}\left(\epsilon^2/L_D\right)$, it can be seen
that $I(D||D'_{KL}) \leq \epsilon$.
\qed
\end{proof}

\begin{corollary}
Consider the problem of learning PDFAs having $n$ states, over
alphabet $\Sigma$, and probabilities represented by bit strings of
length $\ell$. Using sample size (but not time) polynomial in $n$,
$|\Sigma|$ and $\ell$ (and the PAC parameters $\epsilon$ and
$\delta$), a distribution is this class can be estimated within KL
distance $\epsilon$.
\end{corollary}

The proof follows from the observation that such a PDFA can be
represented using a bit string whose length is polynomial in
the parameters.

Consequently we can learn the same class of PDFAs under the
KL-divergence that can be learned under the $L_1$ distance
in~\cite{Palmer:2005}, i.e. PDFAs with distinguishable states but no
restriction on the expected length of their outputs.  However, note
that the hypothesis is ``inefficient'' (a mixture of exponentially
many PDFAs).

\section{Conclusion}

We have shown a close relationship between the error of an estimated
input distribution (as measured by $L_1$ distance or KL-divergence)
and the error rate of the resulting classifier. In situations where
we believe that input distributions may be accurately estimated, the
resulting information about the data may be more useful than just a
near-optimal classifier.

A general issue of interest is the question of when one can obtain
good classifier from estimated distributions that satisfy weaker
goodness-of-approximation criteria than those considered here.
Suppose for example that elements of a 2-element domain $\{x_1,x_2\}$
are being labeled by the stochastic rule that assigns labels 0 and 1
to either element of the domain, with equal probability. Then any
classifier does no better than random labeling, and so we can use
arbitrary distributions $D'_0$ and $D'_1$ as estimates of the
distributions $D_0$ and $D_1$ over examples with label 0 and 1
respectively. In~\cite{Goldberg:2001} we show that in the basic PAC
framework we can sometimes design discriminant functions based on
unlabeled data sets, that result in PAC classifiers without any
guarantee on how well-estimated is the input distribution.
Further work should possibly compromise between the distribution-free
setting, and the objective---considered here---of approximating
the input distributions in a strong sense.

\section{Acknowledgements}

We would like to thank Luc Devroye for drawing to our attention the
statement of the version of
Theorem~\ref{upper-bound_variation_distance} that appears
in~\cite{Devroye}.


\end{document}